\documentclass[journal]{IEEEtran}
\usepackage{amsmath,amsfonts,amssymb}
\usepackage{algorithmic}
\usepackage{algorithm}
\usepackage{array}
\usepackage[caption=false,font=normalsize,labelfont=sf,textfont=sf]{subfig}
\usepackage{textcomp}
\usepackage{stfloats}
\usepackage{url}
\usepackage{verbatim}
\usepackage{graphicx}
\usepackage{cite}
\usepackage{booktabs}
\usepackage{multirow}
\usepackage{float}
\usepackage{orcidlink}
\hypersetup{
    colorlinks=true,
    urlcolor=blue, 
    citecolor=blue, 
    linkcolor=blue
}
\begin{document}

\title{Whether, Which, and Whose: Solving the Triple Challenge of Deepfake Proactive Forensics in Multi-Face Scenarios}

\author{Lei~Zhang\textsuperscript{\orcidlink{0009-0001-6568-1404}},
Zhiqing~Guo\textsuperscript{\orcidlink{0000-0001-6412-334X}},~\IEEEmembership{Member,~IEEE},
Dan~Ma\textsuperscript{\orcidlink{0009-0002-2262-9603}},~\IEEEmembership{Member,~IEEE},

Wenzhong~Yang\textsuperscript{\orcidlink{0000-0002-7017-129X
}},
and~Gaobo~Yang\textsuperscript{\orcidlink{0000-0003-2734-659X}}

\thanks{Lei Zhang, Zhiqing Guo (Corresponding author), Dan Ma, and Wenzhong Yang are with the School of Computer Science and Technology, Xinjiang University, Urumqi, 830017, China. Zhiqing Guo is also with the Xinjiang Multimodal Intelligent Processing and Information Security Engineering Technology Research Center, Urumqi, 830017, China (e-mail:
leiz@stu.xju.edu.cn; \{guozhiqing, madan, yangwenzhong\}@xju.edu.cn).}
\thanks{Gaobo Yang is with College of Computer Science and Electronic Engineering, Hunan University, Changsha, 410082, China. (e-mail: yanggaobo@hnu.edu.cn).}
}

\maketitle

\begin{abstract}
Unlike single-face forgeries, deepfakes in complex multi-person interaction scenarios (such as group photos and multi-person meetings) more closely reflect real-world threats. Although existing proactive forensics solutions demonstrate good performance, they heavily rely on a ``single-face'' setting, making it difficult to effectively deal with the problems of deepfake detection, localization, and source tracing in complex multi-person environments. In this paper, we propose a Deep Attributable Watermarking Framework (DAWF) tailored to multi-face proactive forensics, which establishes an isolated identity attribution space. This spatial isolation ensures that multiple independent tracing signals can coexist within a single image and be successfully anchored to their respective identity instances. Crucially, we propose a selective regional supervision loss to suppress cross-face interference, guiding the decoder to focus exclusively on the manipulated facial regions. DAWF unifies image-level detection, instance-level localization, and identity-level source tracing, successfully achieving the ``whether, which, and whose" forensic goals of determining whether an image is manipulated, which face is forged, and whose identity is tampered with. Extensive experiments on challenging multi-face datasets demonstrate robust triple-forensic performance, achieving an AUC of 0.91, an F1-score of 0.87, and a BER of 0.54\%. The code
is available at \url{https://github.com/vpsg-research/DAWF}.
\end{abstract}  

\section{Introduction}
\begin{figure}[t]
    \centering
    \includegraphics[width=0.45\textwidth]{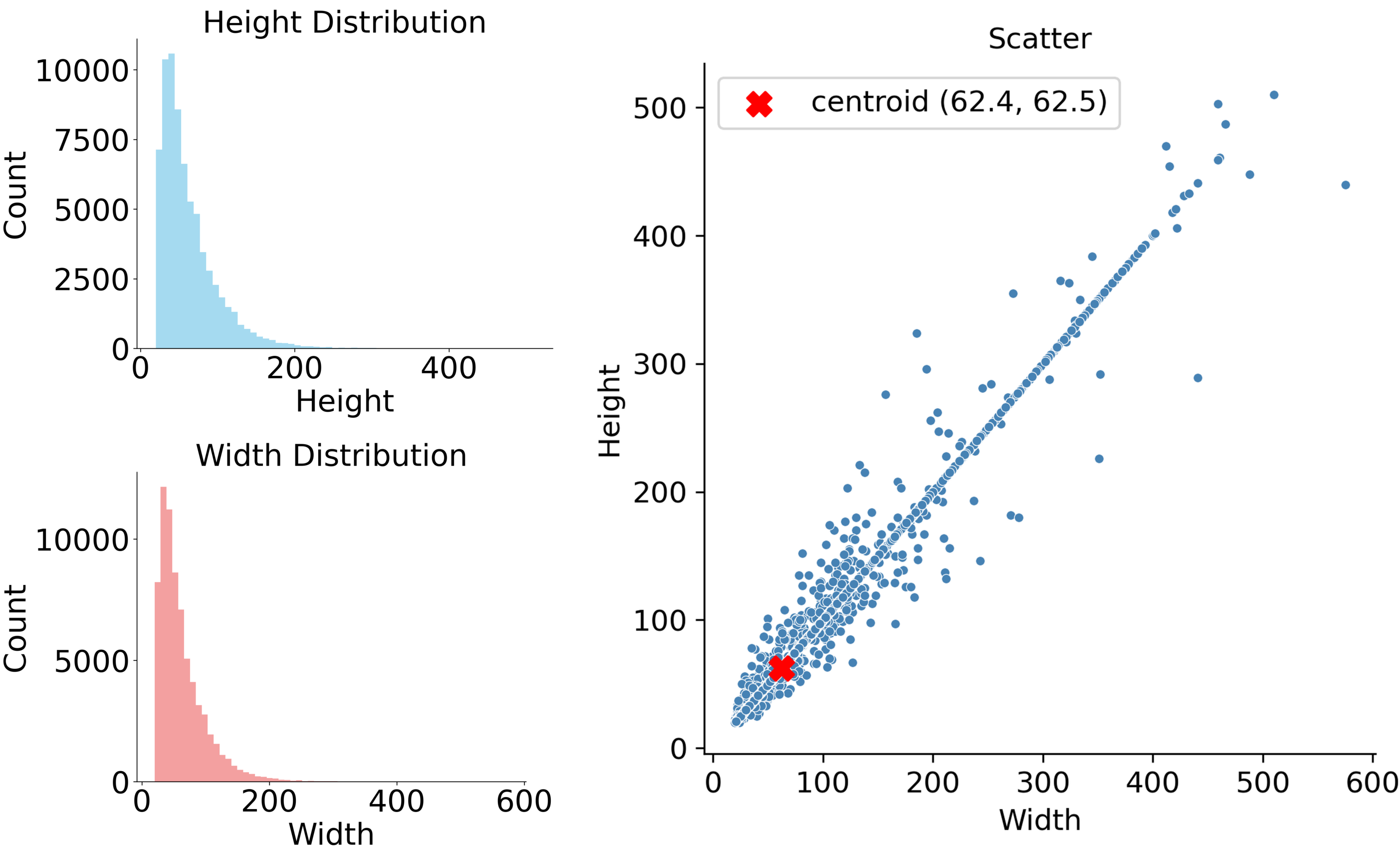}
    \caption{Analysis of detected face resolutions. Left: Histograms showing the frequency distribution of face image heights and widths. Right: Scatter plot displaying the joint distribution of face resolutions. The red ``X'' marker denotes the statistical centroid, centered near $(\mathbf{62.4}, \mathbf{62.5})$.}
    \label{fig:tamper-region2}
\end{figure}

\IEEEPARstart{T}{he} advancement of deepfake technology enables the forgery of complex multi-person interaction scenarios (such as group photos, multi-person meetings, etc.) \cite{zhou2021face7}, \cite{le2021openforensics8}. The seamless blending of forged and authentic identities severely blurs the line between forged content and the real world, making it difficult for common users to distinguish \cite{choi2020stargan}. Extending beyond individual defamation to fabricate social relationships and collective events accelerates the spread of misinformation, cyber fraud, and the erosion of social trust \cite{wang2024deepfake}. Consequently, developing forensics for multi-face deepfakes has become an urgent imperative.

To counter the deepfake threat in multi-face scenarios, passive forensics models the relationship among multiple facial instances to exploit underlying contextual anomalies and inter-face inconsistencies \cite{2li2020sharp17}, \cite{2lin2024exploiting2024}. However, relying purely on black-box detection, these passive approaches inherently struggle to provide a credible chain of evidence, undermining their credibility in both legal contexts and practical deepfake mitigation. In contrast, proactive forensics \cite{wu2026all, 3SUN2025103801S2025D} embeds digital signals prior to content release and extracts them after deepfake manipulation to establish a chain of evidence, ensuring the credibility and reliability of subsequent detection, localization, and source tracing. While recent advances have achieved impressive performance, most proactive forensics methods remain constrained by single-face assumptions (typically one identity per image). To the best of our knowledge, FaceSigns \cite{3neekhara2024facesigns16} is the only early work to extend its framework to multi-face scenarios. However, its evaluation focuses on sparsely distributed and relatively high-resolution faces. When applied to more densely distributed and low-resolution faces, this approach suffers from a severe decline in both watermark robustness and detection performance (detailed in Table \ref{watermark}).

\begin{figure}[t]
    \centering
    \includegraphics[width=0.48\textwidth]{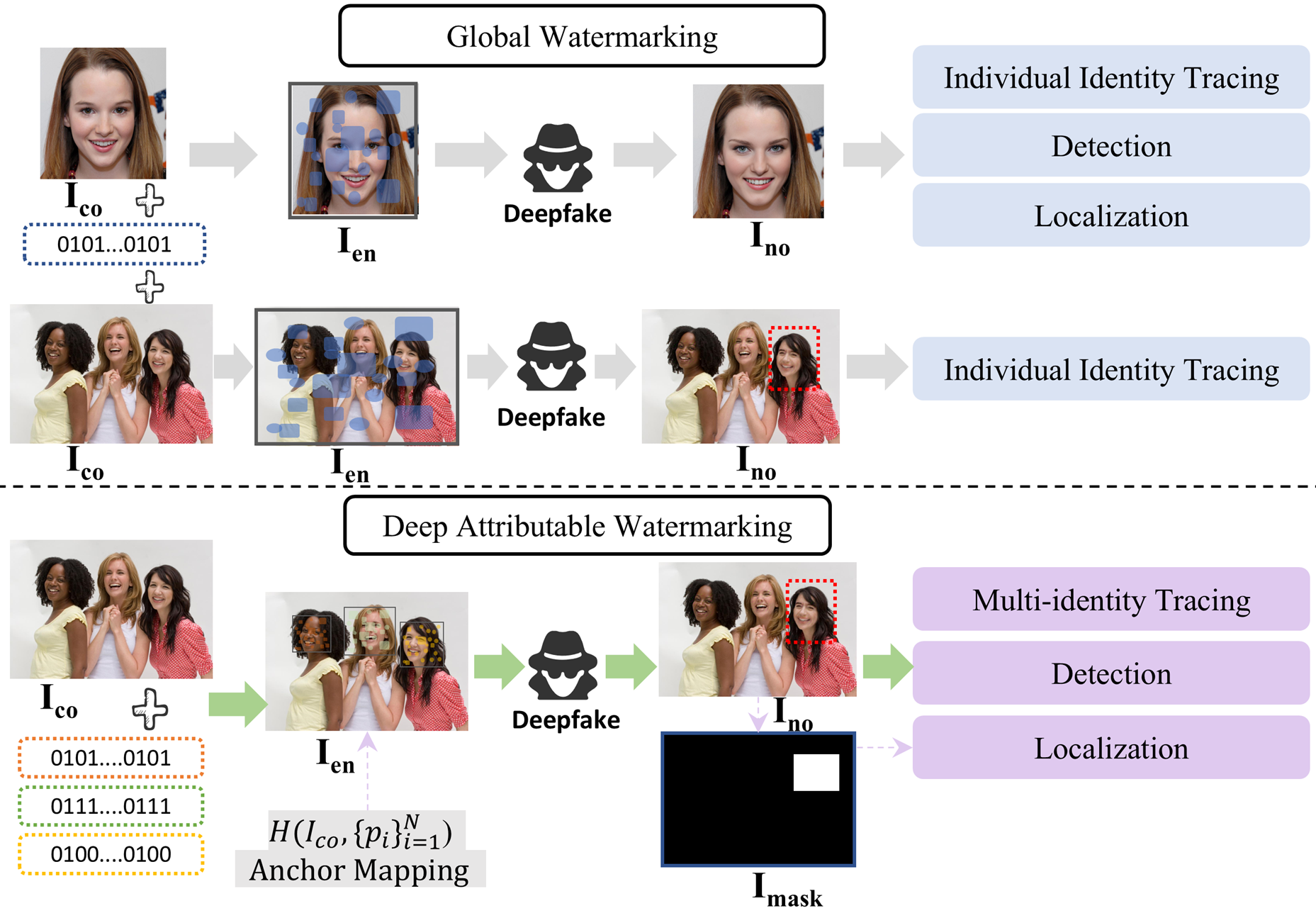}
    \caption{Comparison of proactive forensics methods. While existing global watermarking methods (top) are effective in single-face scenarios, they fail when directly migrated to complex multi-face environments. In contrast, our DAWF (bottom) successfully achieves detection, localization, and multi-identity tracing within a single image.}
    \label{fig:tamper-region1}
\end{figure}

To bridge this critical gap, we elevate proactive deepfake forensics from single-face settings to  complex multi-face scenes, which will face two challenges: (1) Individual faces in multi-person images exhibit significantly lower resolutions compared to single-face images. As shown in Fig. \ref{fig:tamper-region2}, most face instances are concentrated in lower-resolution regions with a statistical centroid near (62.4, 62.5). This sharp decline in spatial resolution causes severe signal dilution within facial areas. Consequently, proactive forensic methods tailored for single-face images exhibit severe sensitivity degradation toward deepfake within multi-face environments, rendering reliable detection and localization unachievable. 
(2) Multi-face environments involve the coexistence of multiple distinct identity instances. As shown in Fig. \ref{fig:tamper-region1}, existing global watermarking methods fail to differentiate among these instances and establish a distinct binding with each identity, rendering multi-identity source tracing impossible. 

To address the aforementioned challenges, we propose the Deep Attributable Watermarking Framework (DAWF). Our framework establishes an isolated identity attribution space to embed a unique watermark into each face, successfully anchoring tracing information to individual facial identities to combat identity theft in multi-face scenarios. Moreover, to mitigate cross-face interference from mixed authentic and fake faces during decoding, we introduce a selective regional supervision loss that actively filters out pristine faces, directing the decoder to focus exclusively on forged regions. Consequently, DAWF answers questions such as ``whether an image is manipulated," ``which face is forged," and ``whose identity is tampered with," establishing a new unified paradigm for multi-identity proactive forensics. The main contributions of our work are summarized as follows:
\begin{itemize}
    \item By proposing the Deep Attributable Watermarking Framework (DAWF), we achieve the ``whether, which, and whose'' forensic goals for the first time in such scenarios, simultaneously detecting manipulated images, identifying tampered faces, and tracing tampered identities.
    
    \item We design an isolated identity attribution space and a selective regional supervision loss. These mechanisms not only anchor unique tracing signals to individual facial instances but also explicitly guide the network to focus on forgery areas, thereby effectively eliminating cross-face interference.

    \item We elevate deepfake proactive forensics from single-face settings to multi-face scenarios, more closely reflecting real-world environments. Extensive experiments on challenging datasets show that DAWF achieves outstanding performance in both accuracy and robustness.
\end{itemize}

The remainder of this paper is organized as follows. Section \ref{sec:related_work} reviews the related work on passive forensics and proactive forensics. Section \ref{sec:method} presents the proposed DAWF framework in detail. Section \ref{sec:experiments} describes the experimental setup and reports the evaluation results. Section \ref{sec:conclusion} concludes the paper.

\section{Related Work}
\label{sec:related_work}

\subsection{Passive Forensics}

Passive forensics techniques achieve detection by analyzing intrinsic anomalies (such as texture distortion and frequency domain noise) and contextual inconsistencies (such as lighting and pose logic) in the forged content. For example, the early representative method SBI \cite{5shiohara2022detecting67} guides the classifier to learn general forgery features by generating ``Self-Blended Images''. Subsequent research has focused on extracting more comprehensive clues by dynamically fusing space-frequency features \cite{guo2023constructing} or integrating local artifacts with global texture information \cite{guo2023ldfnet}. Although these methods excel at single-face detection, their direct application to real-world multi-face scenarios often yields suboptimal results. These approaches fundamentally overlook the crucial informative correlations between faces. To this end, recent methods explicitly model inter-facial relationships through spatio-temporal encoding instances \cite{2li2020sharp17} or the combination of multi-face relationship learning and global feature aggregation \cite{2lin2024exploiting2024} to capture contextual inconsistencies among multiple faces. Furthermore, taking a novel cognitive perspective to handle the complexity of multi-face contexts, Hu et al. \cite{hu2025seeing} introduce a human-inspired framework that simulates human visual observation patterns to enhance the capability of multi-face detection. 

However, these passive methods still rely on black-box detection and cannot trace the specific manipulated faces back to their respective sources in multi-face scenarios. In this paper, by establishing an isolated identity attribution space, we precisely anchor a unique watermark signal to each face instance, shifting from unreliable black-box detection to verifiable detection and successfully achieving multi-identity source tracing.

\subsection{Proactive Forensics}

Proactive forensics embeds a digital watermark into the original image, and then extracts the embedded signal and compares it with the original watermark to achieve detection, localization, and source tracing.  Recent advancements in this field can be systematically categorized across three critical dimensions. First, from the perspective of functional implementation, representative works have integrated source tracing and deepfake detection by introducing deep separable watermarks \cite{3wu2023sepmark18} or modeling proactive forensics as an image-bit steganography problem \cite{zhang2024editguard}. Wu et al. \cite{wu2026all} further propose an all-in-one landmark-identity watermark framework that unifies detection, localization, and source tracing in a single pipeline. He et al. \cite{he2026one} propose Segfacemark, a proactive deepfake detection framework that integrates region-aware watermark embedding with dynamic identity-driven encryption to achieve authenticity verification, forgery classification, and localized forensic interpretability. Second, refining the embedding payload represents another crucial direction, such as embedding identity-specific features to protect unauthorized face-swapping \cite{zhao2023proactive} or leveraging structural priors to construct landmark-based watermarks \cite{6wang2024lampmark25}. Third, from the perspective of emerging security threats and generative models, Jia et al. \cite{6jia2025uncovering} define multi-embedding attacks and propose adversarial interference simulation. Within diffusion models, M{\"u}ller et al. \cite{muller2025black} and Sun et al. \cite{3SUN2025103801S2025D} introduce proactive frameworks to safeguard media against deepfake manipulations. Concurrently, to harmoniously align proactive forensics with these generative workflows, Chen et al. \cite{chen2026forensic} present forensic-friendly controllable latent diffusion to retain manipulability while preserving downstream forensic utility. Notably, Neekhara et al. \cite{3neekhara2024facesigns16} propose FaceSigns, a deep learning-based semifragile watermarking system. While its core networks are trained on single-face data, they extend it to multi-face scenarios through an engineering pipeline that sequentially detects, crops, and resizes individual faces to $256 \times 256$ for watermark embedding before pasting them back into the original image.

However, existing proactive forensics frameworks remain fundamentally confined to single-face settings. Although the pioneering work of FaceSigns has begun to recognize the importance of multi-face forensics, its detection performance severely degrades when confronted with more complex multi-face environments. To bridge this critical gap, we propose DAWF, a proactive forensics framework designed to achieve simultaneous deepfake detection, localization, and source tracing in complex multi-face scenarios.

\section{Methodology}
\label{sec:method}
\begin{figure*}[t]
    \centering
    \includegraphics[width=1.0\textwidth]{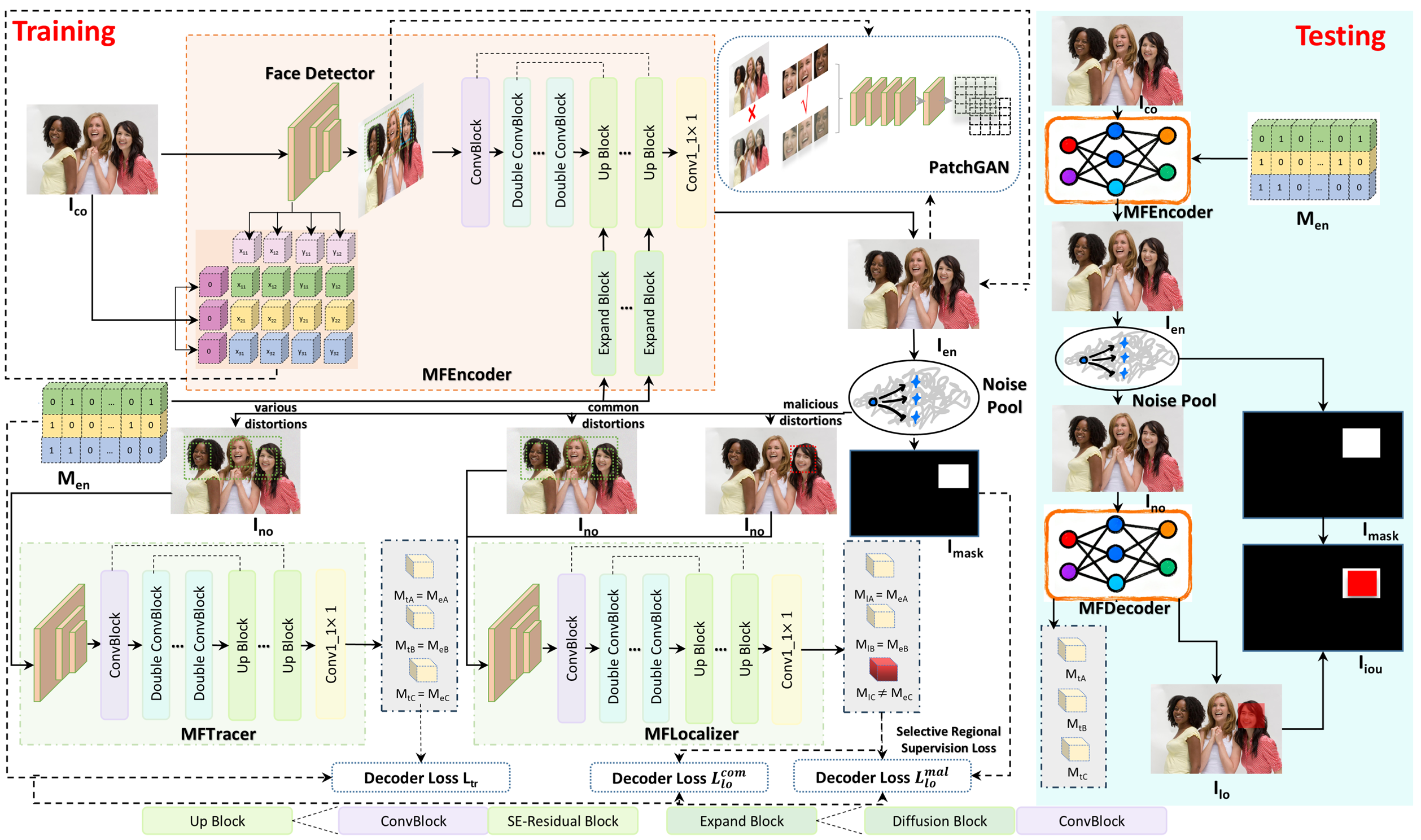}
    \caption{Overall architecture of DAWF. (a) Training phase: The MFEncoder combines a face detector and a steganography kernel to output the encoded image $\mathbf{I}_{\text{en}}$. After undergoing various distortions, the noised image $\mathbf{I}_{\text{no}}$ is passed to the MFTracer and MFLocalizer for watermark recovery, detection, and localization, respectively.
(b) Testing phase: The MFTracer outputs the identity tracing information $\mathbf{M}_{\text{tr}}$, and the MFLocalizer outputs detection result and  localization image $\mathbf{I}_{\text{lo}}$. 
    }
    \label{fig:tamper-region3}
\end{figure*}
\subsection{Overview}
We propose a Deep Attributable Watermarking Framework (DAWF), specifically designed for multi-face scenarios that mirror the intricacies of real-world. As depicted in Fig. \ref{fig:tamper-region3}, the DAWF pipeline is structured into three primary stages. A face detector identifies $N$ facial regions in the cover image $\mathbf{I}_{co}$. The MFEncoder then simultaneously embeds $N$ independent messages into these specific regions, producing the encoded image $\mathbf{I}_{en}$. To simulate various real-world scenarios, $\mathbf{I}_{en}$ is processed through a noise pool to generate the noised images $\mathbf{I}_{no}$. The noise pool are categorized into common perturbations (e.g., JPEG compression) and malicious deepfake manipulations. The MFTracer recovers robust messages from $\mathbf{I}_{no}$ for source attribution. Concurrently, the MFLocalizer detects inconsistencies between embedded and recovered messages to generate a detection result and the localization mask $\mathbf{I}_{lo}$, pinpointing the exact faces modified by deepfake operations.

\subsection{Deep Attributable Watermarking Network}
\textbf{MFEncoder}. Given the cover image $I_{co} \in \mathbb{R}^{3 \times H \times W}$, $N$ facial regions are initially identified. Each face instance $i$ is parameterized by a normalized bounding box vector $v_i = [x_{min}, y_{min}, x_{max}, y_{max}]$. The absolute pixel-level spatial anchors $p_i$ are derived as follows:
\begin{equation}p_i = v_i \odot [W, H, W, H], \quad i = 1, \dots, N\end{equation}
where $\odot$ denotes the element-wise Hadamard product. Based on these coordinates, we establish an isolated identity attribution space for each individual face. Rather than iteratively extracting localized regions, we utilize differentiable bilinear interpolation, denoted as $\mathcal{H}(\cdot)$, to resample all target instances into a fixed $64 \times 64$ resolution simultaneously. This constructs a unified, batch-processed cover tensor $C = \mathcal{H}(I_{co}, \{p_i\}_{i=1}^N) \in \mathbb{R}^{N \times 3 \times 64 \times 64}$, which serves as the precise forensic anchors for parallel embedding. The steganographic kernel jointly processes the cover tensor $C$ and the binary secret messages $M_{en} \in \{-0.1, 0.1\}^{N \times L}$. We introduce a cascaded multi-scale message injection strategy integrated within the U-Net topology. During the expansive phase of the network, the message sequence $M_{en}$ is dynamically projected, reshaped, and spatially resampled to perfectly align with the dimensions of the hierarchical feature maps across multiple scales. These expanded message representations are subsequently fused with the skip-connected image features via dense channel-wise concatenation. This cascaded injection ensures that the watermark signal is adaptively embedded across varying receptive fields, thereby maximizing both extraction robustness and perceptual fidelity. To strictly confine the generated pixel intensities within legal operational boundaries without truncating the gradient propagation flow, the kernel leverages a gradient-preserving differentiable clamping mechanism based on the straight-through estimator (STE). Let $\tilde{S}$ represent the unconstrained encoded patches; the final stego patches $S$ are mathematically formulated as:\begin{equation}S = \tilde{S} + \text{sg}[\Pi_{[-1, 1]}(\tilde{S}) - \tilde{S}]\end{equation}where $\text{sg}[\cdot]$ denotes the stop-gradient operation, and $\Pi_{[-1, 1]}(\cdot)$ represents the projection operator that restricts the values to the $[-1, 1]$ range. Finally, these modulated patches are inversely mapped to their original spatial coordinates to reconstruct the fully encoded image $I_{en}$:
\begin{equation}I_{en} = I_{co} + \sum_{i=1}^{N} \mathcal{H}^{-1}(S_i, p_i)\end{equation}
where $\mathcal{H}^{-1}(\cdot)$ signifies the inverse bilinear interpolation mapping defined by $p_i$.

\textbf{Noise pool}. 
To simulate complex real-world forensic environments, the encoded image $\mathbf{I}_{en}$ is processed by the noise module $\mathcal{N}(\cdot)$ to yield three distinct distorted variants:
\begin{equation}
    \mathbf{I}_{C},\mathbf{I}_{R},\mathbf{I}_{F}=\mathcal{N}(\mathbf{I}_{en})
\end{equation}
The module samples from a comprehensive arbitrary distortion pool, which encompasses both benign perturbations (e.g., compression, blurring) and malicious deepfake manipulations. These three variants are strategically generated to optimize the distinct objectives of our two functional branches: Tracing branch (robust watermark): The variant $\mathbf{I}_{C}$ is generated by subjecting $\mathbf{I}_{en}$ to the arbitrary distortion pool (both benign and malicious). This forces the embedded watermark to remain highly robust against all forms of degradation, ensuring reliable source identity tracing under any condition. Detection and localization branch (semi-fragile watermark): $\mathbf{I}_{R}$ is subjected exclusively to benign perturbations, so that the watermark remains robustly extractable. Conversely, $\mathbf{I}_{F}$ is subjected to malicious deepfake manipulations, where the watermark is intentionally destroyed to yield extraction results approaching random guessing. Furthermore, during the generation of $\mathbf{I}_{F}$, the malicious branch derives the spatial footprint of the deepfake operations to construct a ground-truth mask $\mathbf{I}_{\text{mask}}$. This mask provides critical supervision for our selective regional loss, enabling the MFLocalizer to precisely distinguish maliciously forged regions from pristine or benignly edited areas.

\textbf{MFDecoder}. Given a potentially degraded input $I_{no}$, the network isolates the targeted instances based on $p_i$. A unified extraction kernel processes these localized patches. After aggregating the hierarchical spatial features through symmetrical down-and-up convolutional paths, the terminal feature representations are flattened and projected via linear decoding weights to reconstruct the extracted message vectors $M \in \mathbb{R}^{N \times L}$. To optimize computational efficiency while strictly fulfilling the triple-forensic requirements, the MFTracer and MFLocalizer share identical network topologies but are governed by fundamentally asymmetric optimization constraints. The MFTracer processes $I_{no}$ to robustly reconstruct the source identity information under arbitrary distortions, enforcing the extracted message $M_{tr}$ to closely approximate $M_{en}$. Conversely, the MFLocalizer evaluates signal consistency for precise detection and deepfake localization. It is constrained to reliably recover $M_{lo}^{com}$ in benign scenarios, while being explicitly optimized to yield severely discordant outputs in the presence of malicious alterations. Denoting the decoded outputs as $M_{tr}$, $M_{lo}^{com}$, and $M_{lo}^{mal}$, DAWF enforces the following asymmetric objectives:\begin{equation}M_{tr} \approx M_{en}, \quad M_{lo}^{com} \approx M_{en}, \quad M_{lo}^{mal} \not\approx M_{en}\end{equation} 
This asymmetry is the core mechanism that produces the intended BER polarization.

\textbf{Discriminator $\mathcal{D}$}. During training, the discriminator $\mathcal{D}$ employs the RaLSGAN loss  \cite{8jolicoeur2018relativistic35} exclusively on $64 \times 64$ facial patches. By directly evaluating $\mathbf{C}$ against $\mathbf{S}$, this patch-level discrimination eliminates background interference. It forces the MFEncoder to minimize the distribution gap between pristine and encoded facial textures, strictly confining imperceptibility constraints to target regions for enhanced visual seamlessness.

\textbf{Anchor-aligned mapping.}
\label{sec:Anchor-aligned Mapping}
Our mechanism involves three distinct categories of facial instances: (1) Detected faces ($\mathcal{F}_{det}$), which serve as the forensic anchors where watermarks are actually extracted; (2) Ground-truth forged faces ($\mathcal{F}_{gt}$), representing the actual forged faces; and (3) Predicted forged faces ($\mathcal{F}_{pred}$), which are the instances within $\mathcal{F}_{det}$ identified as ``attacked'' based on their extracted messages. Rather than directly comparing $\mathcal{F}_{gt}$ and $\mathcal{F}_{pred}$, our framework maps both labels onto the detected faces. For each instance $i$ in $\mathcal{F}_{det}$, we assign a binary ground-truth label $y_i$ via IoU matching, and a prediction label $\hat{y}_i$ based on the BER:
\begin{equation}
\begin{aligned}
    y_i &= \begin{cases} 
    1 & \text{if } \exists f \in \mathcal{F}_{gt} \text{ s.t. } \text{IoU}(\mathcal{F}_{det,i}, f) > \tau, \\ 
    0 & \text{otherwise,} 
    \end{cases} \\
    \hat{y}_i &= \begin{cases} 
    1 & \text{if } \text{BER}(\mathcal{F}_{det,i}) > T_{ber}, \\ 
    0 & \text{otherwise.} 
    \end{cases}
\end{aligned}
\end{equation}
This ``anchor-based mapping'' ensures that every extracted watermark is correctly associated with its corresponding facial instance, effectively transforming the localization task into a robust instance-level binary classification. Based on these newly mapped labels $y_i$ and $\hat{y}_i$, we calculate F1-score.
\begin{table}[t]
\caption{Visual quality evaluation of the encoded images.}
\centering
\renewcommand{\arraystretch}{1.15}
\resizebox{\columnwidth}{!}{%
\begin{tabular}{l|c|c|c|c} 
\toprule
Method & Image Size & Message Length & PSNR & SSIM \\ \hline
CIN \cite{3ma2022towards28} & $256 \times 256$ & 128 & 44.6700 & 0.9778 \\ 
SepMark \cite{3wu2023sepmark18} & $256 \times 256$ & 128 & 42.0179 & 0.9759 \\ 
FaceSigns \cite{3neekhara2024facesigns16} & $1024 \times 700$ & $128 \times N$ & \textbf{53.4027} & \textbf{0.9983} \\
WaveGuard \cite{3he2025waveguard2025W}  & $256 \times 256$ & 128 & 45.6446 & 0.9946 \\    
KAD-Net (ST) \cite{6he2025kad} & $256 \times 256$ & 128 & 39.0009 & 0.9569 \\ 
KAD-Net (FD) \cite{6he2025kad} & $256 \times 256$ & 128 & 41.1804 & 0.9671 \\ 
LIDMark \cite{wu2026all} & $256 \times 256$ & 152 & 43.4530 & 0.9801 \\ 
DAWF & $1024 \times 700$ & $15 \times N$ & \underline{51.8010} & \underline{0.9959} \\ 
\bottomrule
\end{tabular}%
}
\label{PSNR}
\end{table}
\begin{figure}[t]
    \centering
    \includegraphics[width=0.47\textwidth]{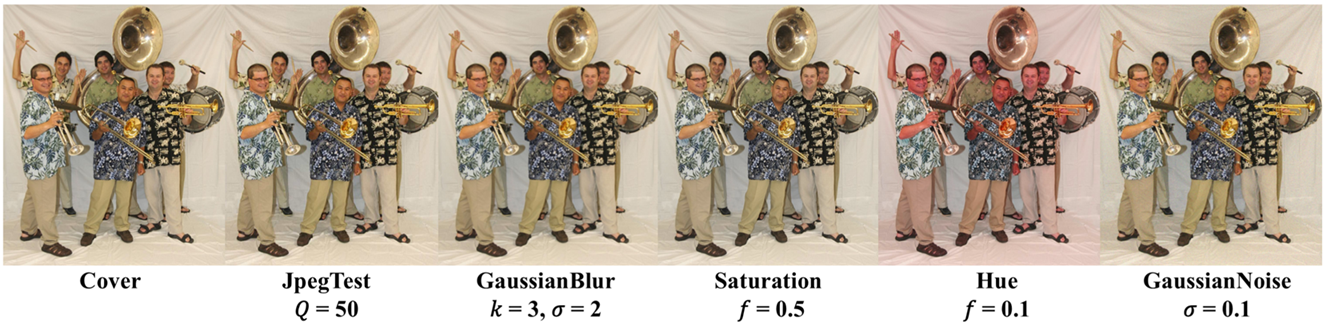}
    \caption{Localization results under common noise attacks (zero false alarms).}
    \label{common}
\end{figure}

\subsection{Loss Functions}
We jointly optimize the MFEncoder and the MFDecoder by minimizing a comprehensive loss function $\mathcal{L}_{total}$, while updating the discriminator $\mathcal{D}$ in an adversarial manner. This joint optimization is driven by two primary objectives: the visual imperceptibility of the encoded images and the triple-forensic accuracy of message recovery.
\begin{figure}[t]
    \centering
    \includegraphics[width=0.5\textwidth]{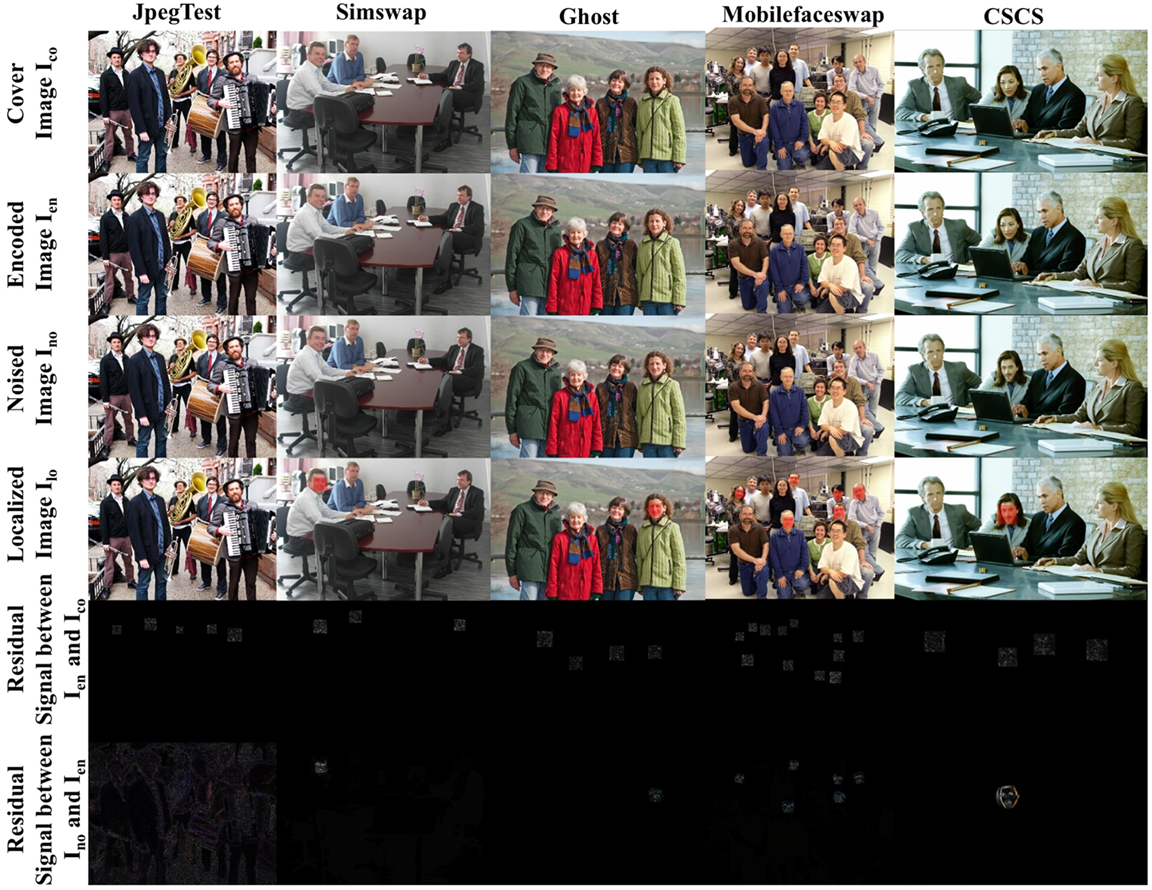}
    \caption{Localization results and visual quality under various typical distortions. Each column represents a distinct distortion type, with localization shown by the red overlay. The examples illustrate that DAWF preserves visual fidelity while retaining high accuracy in deepfake localization.}
    \label{visiual}
\end{figure}

\begin{table*}[t]
\centering
\caption{Quantitative comparison on WIDERFace dataset regarding BER (\%) of the watermarks under various distortions.}
\label{watermark}
\renewcommand{\arraystretch}{1.15}
\resizebox{\textwidth}{!}{%
\begin{tabular}{l|c c|c c|c c|c c|c c|c c|c c}
\toprule
\multirow{2}{*}{Distortion} & 
\multicolumn{2}{c|}{CIN \cite{3ma2022towards28}} & 
\multicolumn{2}{c|}{SepMark \cite{3wu2023sepmark18}} & 
\multicolumn{2}{c|}{FaceSigns \cite{3neekhara2024facesigns16}} & 
\multicolumn{2}{c|}{WaveGuard \cite{3he2025waveguard2025W}} & 
\multicolumn{2}{c|}{KAD-Net \cite{6he2025kad}} & 
\multicolumn{2}{c|}{LIDMark \cite{wu2026all}} & 
\multicolumn{2}{c}{DAWF} \\
& Tracer & Detector & Tracer & Detector & Tracer & Detector & Tracer & Detector & Tracer & Detector & Tracer & Detector & Tracer & Localizer \\ \hline
Jpeg & 49.5966 & - & 9.2780 & 12.5729 & - & 49.8314 & 0.0002 & 0.0004 & 7.6242 & 8.5365 &1.5432&- & 0.9154 & 0.9957 \\ 
GaussianBlur & 3.5495 & - & 0.7328 & 12.2367 & - & 43.1008 & 0.0000 & 0.0000 & 0.0690 & 4.0102 &0.0040&- & 0.1777 & 0.4206 \\ 
Saturation & 3.5188 & - & 0.0015 & 34.1703 & - & 8.9372  & 0.0000 & 0.0000 & 0.0000 & 35.0292 &0.2751&- & 0.0071 & 0.2099 \\ 
Hue & 3.5217 & - & 0.0073 & 27.6727 & - & 13.9944 & 0.0004 & 0.0002 & 0.0000 & 32.5529 &0.1082&- & 0.0099 & 0.2171 \\ 
GaussianNoise & 3.6407 & - & 9.9924 & 13.5305 & - & 47.4763 & 0.0298 & 0.0332 & 3.8836 & 6.4875 &0.9372&- & 1.1493 & 1.3118 \\ 
\hline
Average & 12.7655 & - & 4.0024 & 20.0366 & - & 32.6680 & \textbf{0.0061} & 0.0068 & 2.3154 & 17.3233 &0.5735&- & \underline{0.4518} & \textbf{0.6310} \\ 
\hline
Simswapmulti & 3.5497 & - & 0.1697 & 27.4774 & - & 34.6664 & 0.0015 & 0.0021 & 1.0908 & 28.1003 &0.8463&- & 0.5207 & 36.6685\\ 
Ghost & 3.5431 & - & 0.0020 & 36.3050 & - & 35.2920 & 0.0015 & 0.0019 & 0.1315 & 28.9776 &0.1271&- & 0.0490 & 37.4588 \\ 
Mobilefaceswap & 3.5091 & - & 0.0746 & 26.7088 & - & 40.6305 & 0.0015 & 0.0021 & 0.0003 & 27.0410 &0.3509&- & 0.1088 & 43.7540 \\ 
CSCS & 3.5936 & - & 2.0700 & 31.8544 & - & 36.5642 & 0.0015 & 0.0021 & 2.9845 & 29.0502 &0.5056&- & 0.4050 & 44.8044 \\ 
\hline
Average & 3.5488 & - & 0.5790 & 30.5864 & - & 36.7882 & \textbf{0.0015} & 0.0021 & 1.0517 & 28.2922 &0.4575&- & \underline{0.2708} & \textbf{40.6714} \\ 
\bottomrule
\end{tabular}%
}
\end{table*}
\begin{table*}[t]
\centering
\caption{Quantitative comparison with tamper localization methods under malicious distortions.}
\label{temper}
\renewcommand{\arraystretch}{1.15}
\resizebox{\textwidth}{!}{%
\begin{tabular}{l|c c c|c c c|c c c|c c c}
\toprule
\multirow{2}{*}{Method} & \multicolumn{3}{c|}{Simswap} & \multicolumn{3}{c|}{Ghost} & \multicolumn{3}{c|}{Mobilefaceswap} & \multicolumn{3}{c}{CSCS} \\
& F1 & AUC & $BER_{tr}$(\%) & F1 & AUC & $BER_{tr}$(\%) & F1 & AUC & $BER_{tr}$(\%) & F1 & AUC & $BER_{tr}$(\%) \\ \hline
MVSS-Net \cite{dong2022mvss} & 0.2656 & 0.6391 & - & 0.2658 & 0.6446 & - & 0.2351 & 0.6404 & - & 0.2695 & 0.6586 & - \\ 
IML-VIT \cite{ma2023iml} & 0.1661 & 0.4600 & - & 0.1619 & 0.4704 & - & 0.1045 & 0.4304 & - & 0.1679 & 0.4712 & - \\ 
EditGuard \cite{zhang2024editguard} & 0.5954 & 0.7401 & 0.7599 & 0.5632 & 0.7082 & 0.4504 & 0.5906 & 0.7953 & 0.7501 & 0.5053 & 0.7368 & 0.5540 \\ 
OmniGuard \cite{zhang2025omniguard} & 0.0000 & 0.5000 & 0.8676  & 0.0000 & 0.5000 & 0.6423  & 0.0000 & 0.5000 & 0.9496 & 0.0000 & 0.5000 & 0.7843 \\ 
PIM \cite{kong2025pixel} & 0.2036 & 0.5843 & -  & 0.1945 & 0.5822 & -  & 0.2032 & 0.5253 & - & 0.2025 & 0.5562 & - \\ 
DAWF & \textbf{0.7985} & \textbf{0.8722} & \textbf{0.5207} & \textbf{0.8772} & \textbf{0.9108} & \textbf{0.0490} & \textbf{0.8544} & \textbf{0.9333} & \textbf{0.1088} & \textbf{0.9368} & \textbf{0.9403} & \textbf{0.4050} \\
\bottomrule
\end{tabular}%
}
\end{table*}

We constrain the MFEncoder's embedding intensity using the squared $L_2$ distance between $\mathbf{C}$ and $\mathbf{S}$:
\begin{equation}
\mathcal{L}_{en} = \big\| \mathbf{S} - \mathbf{C} \big\|_2^2
\end{equation}
Employing the RaLSGAN formulation, the objective for $\mathcal{D}$ is to maximize the relative margin between the  cover patches and the encoded patches:
\begin{equation}
\begin{aligned}
\mathcal{L}_{D} =\;& \mathbb{E}_{\mathbf{C}}\big(\big(D(\mathbf{C}) - \mathbb{E}_{\mathbf{S}}(D(\mathbf{S})) + 1\big)^2\big) \\
&+ \mathbb{E}_{\mathbf{S}}\big(\big(D(\mathbf{S}) - \mathbb{E}_{\mathbf{C}}(D(\mathbf{C})) - 1\big)^2\big)
\end{aligned}
\end{equation}
Conversely, the adversarial loss guides MFEncoder to deceive $\mathcal{D}$:
\begin{equation}
    \begin{aligned}
    \mathcal{L}_{adv} =\;& \mathbb{E}_{\mathbf{C}}\big(\big(D(\mathbf{C}) - \mathbb{E}_{\mathbf{S}}(D(\mathbf{S})) - 1\big)^2\big) \\
&+ \mathbb{E}_{\mathbf{S}}\big(\big(D(\mathbf{S}) - \mathbb{E}_{\mathbf{C}}(D(\mathbf{C})) + 1\big)^2\big)
\end{aligned}
\end{equation}
To enable the framework to respond dynamically to different types of image degradations, we compute distinct loss components based on the outputs of the MFTracer and MFLocalizer branches:
\begin{equation}
\mathcal{L}_{tr} = \big\| \mathbf{M}_{tr} - \mathbf{M}_{en} \big\|_2^2
\end{equation}
\begin{equation}
\mathcal{L}_{lo}^{com} = \big\| \mathbf{M}_{lo}^{com} - \mathbf{M}_{en} \big\|_2^2
\end{equation}
Here, $\mathcal{L}_{tr}$ ensures the robustness of the source attribution under arbitrary distortions, while $\mathcal{L}_{lo}^{com}$ enforces message consistency under common perturbations.

To achieve extreme sensitivity to malicious deepfake manipulations, we propose a selective regional supervision loss $\mathcal{L}_{lo}^{mal}$. This loss is exclusively applied to facial regions that have been altered by deepfake, driving their decoded messages toward a predefined ``failure state'' (a zero vector $\mathbf{0}$). To formally define the affected regions, let $\mathcal{B}_{face} = \{\mathbf{p}_i\}_{i=1}^N$ be the set of watermarked face bounding boxes, and let $\mathcal{B}_{fake} = \{\mathbf{f}_j\}_{j=1}^K$ be the set of forged regions derived from the ground-truth mask. We construct the target set $\Omega_{fake}$ containing indices of maliciously forged faces by evaluating the Intersection over Union (IoU) \cite{7zheng2020distance8} :
\begin{equation}
\begin{aligned}
\Omega_{fake} = \big\{ & i \in \{1, \dots, N\} \;\big|\; \\
& \exists \mathbf{f}_j \in \mathcal{B}_{fake} \text{ s.t. } \text{IoU}(\mathbf{p}_i, \mathbf{f}_j) > \tau \big\}
\end{aligned}
\end{equation}
where $\tau$ is a predefined overlap threshold. $\mathcal{L}_{lo}^{mal}$ is then strictly calculated over this marked set $\Omega_{fake}$:
\begin{equation}\mathcal{L}_{lo}^{mal} = \frac{1}{|\Omega_{fake}|} \sum_{i \in \Omega_{fake}} \big\| \mathbf{M}_{lo, i}^{mal} - \mathbf{0} \big\|_2^2
\end{equation}
Finally, the total loss $\mathcal{L}_{total}$ combining all optimization objectives is formulated as:
\begin{equation}\mathcal{L}_{total} = \lambda_1 \mathcal{L}_{adv} + \lambda_2 \mathcal{L}_{en} + \lambda_3 \mathcal{L}_{tr} + \lambda_4 \mathcal{L}_{lo}^{com} + \lambda_5 \mathcal{L}_{lo}^{mal}
\end{equation}
where $\lambda_1, \dots, \lambda_5$ are configurable weighting coefficients designed to balance visual imperceptibility and triple-forensic recovery performance.

\section{Experiments}
\label{sec:experiments}
\subsection{Implementation Details}

We employ the WIDERFace dataset \cite{7yang2016wider1}, a widely used and challenging  multi-face detection benchmark in computer vision, to evaluate the performance of DAWF. The images are organized into 61 real-world event categories, such as group photos, sporting events, and dinner parties, providing diverse scene backgrounds and face distributions. To match the operational requirements of current face-swapping pipelines, we construct a filtered subset of WIDERFace by retaining facial instances with bounding boxes larger than $20 \times 20$ pixels, since smaller instances usually contain insufficient structural detail for reliable face swapping and forensic evaluation. This choice is further motivated by prior face-detection studies showing that faces below 20 pixels fall into a particularly challenging tiny-face regime \cite{hu2017finding}, making them unsuitable for high-fidelity face swapping. The training, validation, and testing splits strictly follow the original partitioning of WIDERFace, with the same filtering criterion applied within each split.
   
Our DAWF is implemented in PyTorch \cite{7paszke2019pytorch7} and executed on an NVIDIA A40 GPU. The model is optimized on the filtered WIDERFace subset using the Adam optimizer \cite{7kingma2014adam6} for 100 epochs, with a learning rate of $4 \times 10^{-4}$ and a batch size of 64. Specifically, the detected facial regions are resampled to $64 \times 64$ as input to the steganographic kernel. An independent 15-bit message is embedded into each face, yielding a total watermark capacity of $15 \times N$ bits for a $1024 \times 700$ cover image, where $N$ denotes the number of detected faces. PSNR, SSIM, and BER are used to evaluate visual quality and watermark robustness, while AUC and F1-score are used to assess deepfake detection and localization performance. Since existing frameworks are primarily designed for single-face scenarios, no prior method directly addresses the unified forensic task of simultaneous deepfake detection, localization, and source tracing in multi-face environments. Therefore, we conduct separate comparisons with state-of-the-art proactive forensics and tamper localization methods. All competing methods are evaluated on the same filtered subset of WIDERFace.  
\begin{figure*}[htbp]
    \centering
    \includegraphics[width=1.0\textwidth]{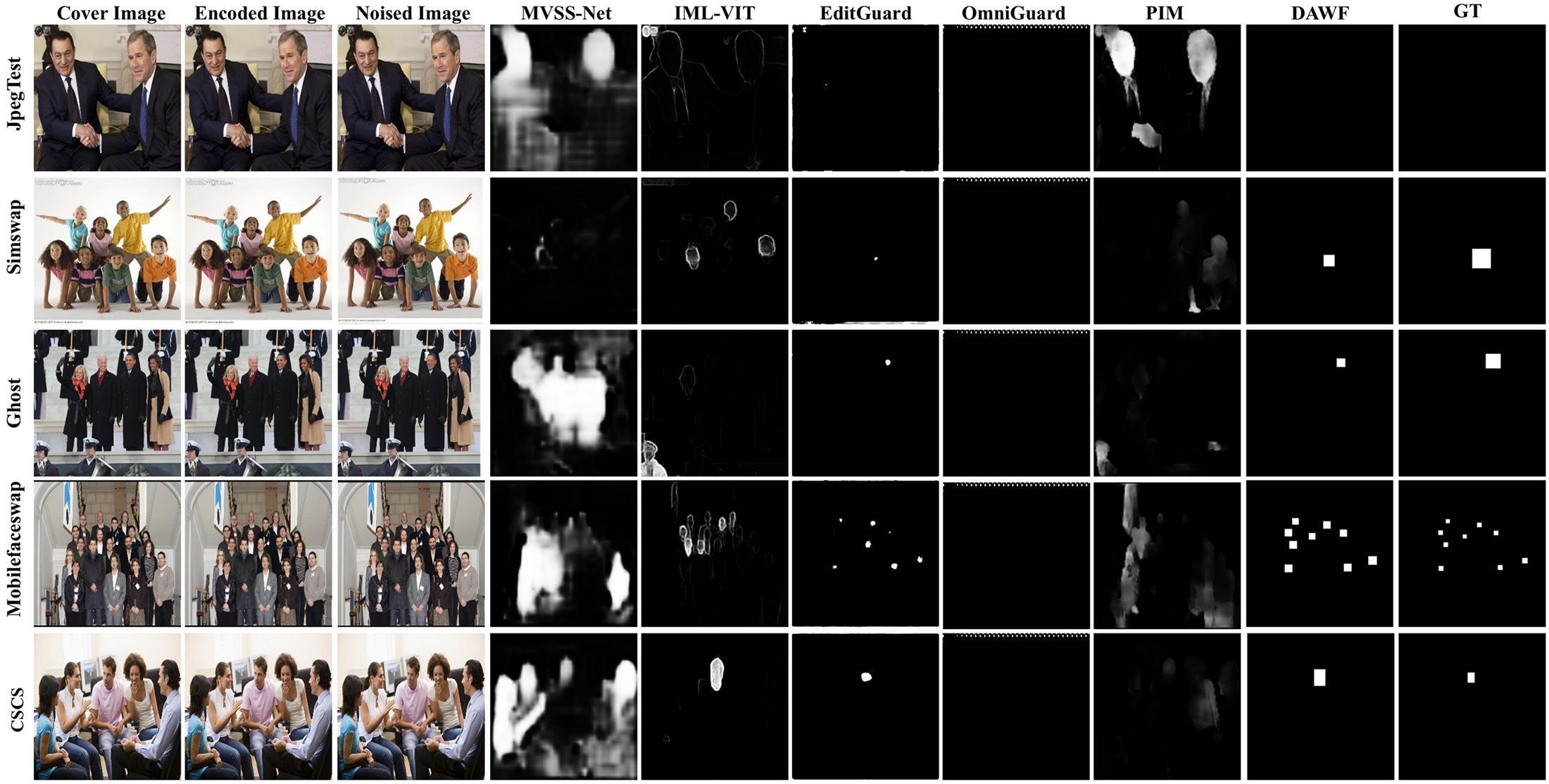}
    \caption{Qualitative comparison of deepfake localization on the WIDERFace dataset. Passive baselines generate extensive false alarms under benign JPEG compression, while competing proactive methods produce ambiguous scattered artifacts (or blurry outlines) and fail to isolate the targets in mixed multi-face scenarios. In contrast, DAWF cleanly highlights only the forged faces, demonstrating its superior capability without being interfered by authentic faces or benign edits.}
    \label{localization}
\end{figure*}
\begin{figure}[t]
    \centering
    \includegraphics[width=0.47\textwidth]{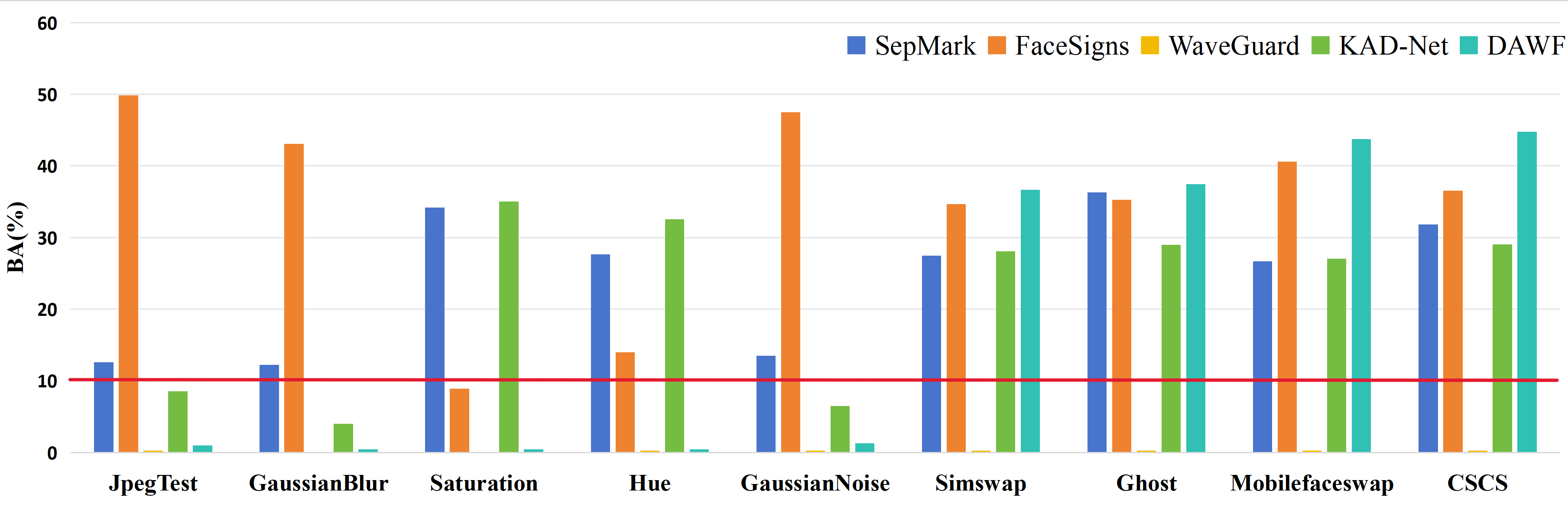}
    \caption{Comparison of the detector's BER under various attacks. Only our DAWF model demonstrates a clear boundary (indicated by the red line) between common noises and deepfake manipulations, proving its capability for deepfake detection and localization.}
    \label{ber}
\end{figure}

To comprehensively evaluate deepfake detection and localization capability, we employ four representative face-swapping models: (1) SimSwap \cite{chen2020simswap}, an identity-agnostic framework that achieves arbitrary identity transfer while preserving target attributes via weak feature matching; (2) Ghost \cite{7groshev2022ghost4}, an attention-driven generative model renowned for high-fidelity synthesis and seamless blending of skin tones and complex expressions; (3) MobileFaceSwap \cite{7xu2022mobilefaceswap5}, a highly efficient model optimized for mobile deployment using dynamic neural network techniques, introducing unique lightweight degradation patterns typical of real-world, low-resource deepfake scenarios; and (4) CSCS \cite{huang2024identity}, a framework utilizing dual surrogate generative models for explicit identity supervision. Notably, since the original CSCS architecture is limited to single-face swapping, we customized it by integrating face detection and seamless blending modules, thereby enabling us to rigorously test our DAWF in multi-face forensic environments.

\subsection{Comparison with Proactive Forensics Methods}
\label{sec:pro_comwork}
DAWF pursues two distinct goals: robustness and semi-fragility. Source tracing (Tracer): The goal is strictly robust identity recovery. Therefore, the expected BER should consistently approach zero under all conditions. Forgery detection and localization (Detector/Localizer): The goal is a semi-fragile watermark. The watermark must survive benign operations but explicitly ``break" when encountering malicious deepfake manipulations. Consequently, the Localizer requires a low BER under common perturbations, but a high BER under malicious attacks to highlight the forged regions. 
\begin{figure}[t]
    \centering
    \includegraphics[width=0.48\textwidth]{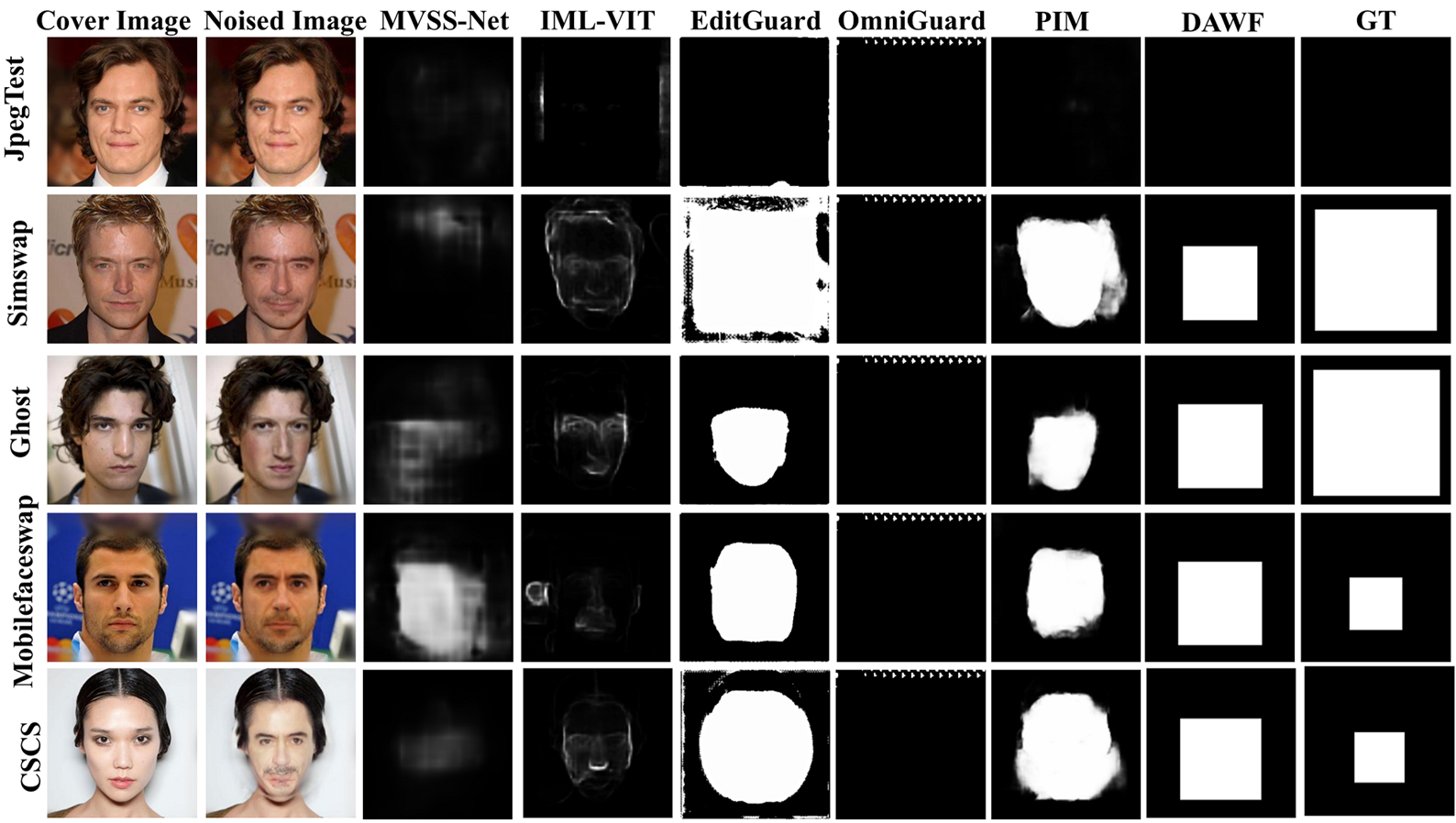}
    \caption{Localization precision comparisons of DAWF and competing methods on CelebA-HQ. The successful application of DAWF to single-face scenarios further validates its broad applicability.}
    \label{celebahq}
\end{figure}

\textbf{Visual quality and source tracing robustness.} We evaluate the visual quality of encoded images using PSNR and SSIM as primary objective metrics. It is worth noting that most baseline methods are evaluated at a resolution of $256 \times 256$. Since these methods distribute watermark signals globally, deploying them in high-resolution scenarios would incur prohibitive computational overhead. Therefore, we evaluate these baselines adhering strictly to the native dimensions specified in their original papers. As shown in Table \ref{PSNR}, DAWF clearly outperforms KAD-Net \cite{6he2025kad} and SepMark \cite{3wu2023sepmark18} in visual fidelity, while performing comparably to FaceSigns \cite{3neekhara2024facesigns16}. Although FaceSigns \cite{3neekhara2024facesigns16} achieves the top PSNR, its detector branch in Table \ref{watermark} reports very high average BER under both common perturbations (32.6680\%) and malicious attacks (36.7882\%), meaning the detector is functionally unstable and fails to provide reliable detection. Regarding source tracing robustness, as reported in Table \ref{watermark}, this tracing robustness is highly competitive, exhibiting a marginal gap of only 0.7150\% behind state-of-the-art methods like WaveGuard. Although WaveGuard \cite{3he2025waveguard2025W} achieves the lowest tracing BER, its detector BER remains near zero under both common and malicious distortions (0.0068\% and 0.0021\%, respectively), indicating near-complete insensitivity to forged-region changes. Unlike these existing methods restricted to image-level or single-identity tracing, DAWF efficiently achieves multi-identity source tracing for multiple distinct identities within the same image. It efficiently hides a dynamic $15 \times N$-bit payload and naturally adapts to the number of faces. This significantly enhances flexibility for complex multi-person scenarios. Therefore, although FaceSigns and WaveGuard each excel in a single metric, both lose a core detection capability required by proactive deepfake forensics. In contrast, by successfully reconciling these critical trade-offs, DAWF achieves superior overall forensic performance for proactive forensics in complex multi-face environments. As illustrated in Fig. \ref{visiual}, our framework maintains high visual fidelity while preserving accurate deepfake localization.

\begin{table*}[t]
\centering
\caption{Quantitative comparison on different datasets under malicious distortions.}
\label{dataset}
\renewcommand{\arraystretch}{1.15}
\resizebox{\textwidth}{!}{%
\begin{tabular}{l|c c c|c c c|c c c}
\toprule
\multirow{2}{*}{Distortion} & \multicolumn{3}{c|}{COCO2017 \cite{7lin2014microsoft2}} & \multicolumn{3}{c|}{OpenForensics \cite{le2021openforensics8}} & \multicolumn{3}{c}{CelebA-HQ \cite{karras2017progressive}} \\ 
 & F1 & AUC & $BER_{tr}$(\%) & F1 & AUC & $BER_{tr}$(\%) & F1 & AUC & $BER_{tr}$(\%) \\ \hline
Simswap & 0.7363 & 0.8350 & 0.5387 & 0.8724 & 0.7826 & 0.6478 & 0.8686 & 0.9436 & 0.3307 \\ 
Ghost & 0.8552 & 0.8937 & 0.0778 & 0.9130 & 0.8043 & 0.2065 & 0.8469 & 0.9324 & 0.0463 \\ 
Mobilefaceswap & 0.8279 & 0.9255 & 0.0704 & 0.9020 & 0.8261 & 0.2307 & 0.8701 & 0.9452 & 0.0412 \\
CSCS & 0.9061 & 0.9190 & 0.4436 & 0.9449 & 0.8261 & 0.4183 & 0.8732 & 0.9460 & 0.1597 \\ \bottomrule 
\end{tabular}
}
\end{table*}

\begin{table*}[t]
\centering
\caption{Localization and $\text{BER}_{tr}$ performance of our DAWF and IML-Net \cite{ma2023iml} under different levels of distortions.}
\label{tab:robustness_merged}
\renewcommand{\arraystretch}{1.1}
\resizebox{\textwidth}{!}{%
\begin{tabular}{l|l|c|c c|c c c|c c c c}
\toprule
\multirow{2}{*}{Methods} & \multirow{2}{*}{Metrics} & \multirow{2}{*}{Clean} & \multicolumn{2}{c|}{Saturation} & \multicolumn{3}{c|}{GaussianBlur} & \multicolumn{4}{c}{JPEG} \\ 
 & & & $f=0.2$ & $f=0.3$ & $k=3,\sigma=0.5$ & $k=3,\sigma=1$ & $k=5,\sigma=1.5$ & 60 & 70 & 80 & 90 \\ \hline
\multirow{2}{*}{IML-ViT \cite{ma2023iml}} & F1 & 0.0661 & 0.0626 & 0.0626 & 0.0578 & 0.0454 & 0.0412 & 0.0469 & 0.0490 & 0.0525 & 0.0633 \\
 & $BER_{tr}$ (\%) & - & - & - & - & - & - & - & - & - & - \\ \hline
\multirow{2}{*}{DAWF} & F1 & 0.7985 & 0.7941 & 0.7940 & 0.7933 & 0.7925 & 0.7913 & 0.7971 & 0.7972 & 0.7975 & 0.7982 \\
 & $BER_{tr}$ (\%) & 0.5207 & 0.5526 & 0.5676 & 0.5570 & 0.5973 & 1.0079 & 2.6844 & 2.3929 & 2.2152 & 2.0123 \\
\bottomrule
\end{tabular}%
}
\end{table*}

\begin{table}[t]
\caption{Quantitative experiments on average watermark embedding and extraction times.}
\centering
\renewcommand{\arraystretch}{1.15}
\setlength{\tabcolsep}{9pt}
\begin{tabular}{l|c|c} 
\toprule
Method & Embed Time (s)  & Extract Time (s) \\ \hline
CIN \cite{3ma2022towards28} & 0.0193 & 0.0182 \\ 
SepMark \cite{3wu2023sepmark18} & 0.0112 & 0.0170 \\ 
FaceSigns \cite{3neekhara2024facesigns16} & 0.0191 & 0.0197 \\ 
WaveGuard \cite{3he2025waveguard2025W} & 0.0714 & 0.0706 \\ 
KAD-Net (ST) \cite{6he2025kad} & 0.0105 & 0.0156 \\ 
KAD-Net (FD) \cite{6he2025kad} & 0.0114 & 0.0177 \\ 
LIDMark \cite{wu2026all} & 0.0156 & 0.0184 \\ \hline
DAWF & 0.0131 & 0.0188 \\ 
\bottomrule
\end{tabular}
\label{tab:model_time}
\end{table}

\textbf{Distinctive deepfake detection and localization capability.} Driven by our selective regional supervision loss, DAWF establishes a clear decision boundary. As illustrated in Fig. \ref{ber}, setting a unified BER threshold (e.g., 10\%) reveals the significant limitations of existing methods. For instance, SepMark \cite{3wu2023sepmark18} and KAD-Net \cite{6he2025kad} show high BERs under common distortions, suggesting that their detectors sacrifice robustness against benign edits to maintain high sensitivity to deepfake operations. Furthermore, LIDMark \cite{wu2026all} relies on 68 facial landmarks extracted from a single face for its detection and localization mechanisms. This design is inherently incompatible with complex multi-face scenarios. Consequently, we primarily restrict the evaluation of LIDMark to its global source tracing capability. In contrast, DAWF successfully maintains a distinct decision boundary, validating its excellent ability to achieve deepfake detection and localization. As visually demonstrated in Fig. \ref{common}, no false alarms are generated across various common noise attacks.

\subsection{Comparison with Tamper Localization Methods}
DAWF specifically focuses on instance-level localization. When analyzing an image containing a mixture of authentic and forged faces, the primary requirement is twofold: reliably localizing only the manipulated faces, and ensuring no false alarms are triggered by benign perturbations. As reported in Table \ref{temper}, DAWF consistently outperforms all competing baselines by significant margins in both detection and localization, achieving an average AUC of 0.9142 and an F1-score of 0.8667 under malicious distortions. Notably, against the CSCS attack, DAWF achieves a remarkable F1-score of 0.9368. This superiority is further corroborated by the qualitative results in Fig. \ref{localization}: passive models exhibit severe over-sensitivity to benign distortions like JPEG compression, leading to false positives, while other proactive methods struggle to localize forged faces in multi-face scenarios. Furthermore, while maintaining this superior detection and localization performance, our framework ensures a near-zero tracing BER (e.g., 0.0490\% under Ghost), whereas all other methods fail to achieve effective triple forensics in complex multi-face environments.

\begin{table*}[htbp]
\centering
\caption{Ablation studies on the core components of DAWF.}
\label{tab:ablation_study}
\renewcommand{\arraystretch}{1.15} 
\resizebox{\textwidth}{!}{%
\begin{tabular}{c c c|c|c|c c|c c|c c}
\toprule
\multirow{2}{*}{$\mathcal{L}_{tr}$} & \multirow{2}{*}{$\mathcal{L}_{lo}^{com}$} & \multirow{2}{*}{$\mathcal{L}_{lo}^{mal}$} & \multirow{2}{*}{PSNR} & \multirow{2}{*}{SSIM} & \multirow{2}{*}{F1} & \multirow{2}{*}{AUC} & \multicolumn{2}{c|}{Tracer} & \multicolumn{2}{c}{Localizer} \\ 
 & & & & & & & $\text{BER}_{tr}^{com}$(\%) & $\text{BER}_{tr}^{mal}$(\%) & $\text{BER}_{lo}^{com}$(\%) & $\text{BER}_{lo}^{mal}$(\%) \\ \hline
0.0 & \checkmark & \checkmark & 53.0595 & 0.9968 & 0.6987 & 0.7847 & 50.0516 & 50.1159 & 5.0058 & 23.5251 \\ 
\checkmark & 0.0 & \checkmark & 45.7501 & 0.9919 & 0.5090 & 0.5000 & 27.5746 & 20.9319 & 50.0938 & 50.1069 \\ 
\checkmark & \checkmark & 0.0 & 47.6685 & 0.9941 & 0.0007 & 0.5003 & 2.0810  & 0.0046 & 1.8843 & 0.0089 \\ 
\checkmark & \checkmark & \checkmark & 51.8010 & 0.9959 & 0.8667 & 0.9142 & 0.4518 & 0.2708 & 0.6310 & 40.6714 \\
\bottomrule
\end{tabular}%
}
\end{table*}
\begin{table*}[t]
    \centering
    \caption{Ablation experiments: Performance of DAWF under varying operational resolutions and payload capacities.}
    \label{tab:performance_metrics}
    \renewcommand{\arraystretch}{1.15}
    \setlength{\tabcolsep}{8pt}
    \begin{tabular}{c|c|c|c|c|c|c c|c c}
        \toprule
        \multirow{2}{*}{Size} & \multirow{2}{*}{Message Length} & \multirow{2}{*}{PSNR} & \multirow{2}{*}{SSIM} & \multirow{2}{*}{F1} & \multirow{2}{*}{AUC} & \multicolumn{2}{c|}{Tracer} & \multicolumn{2}{c}{Localizer} \\ 
        & & & & & & $\text{BER}_{tr}^{com}$(\%) & $\text{BER}_{tr}^{mal}$(\%) & $\text{BER}_{lo}^{com}$(\%) & $\text{BER}_{lo}^{mal}$(\%) \\ \hline
        $64 \times 64$ & $15 \times N$ & 51.8010 & 0.9959 & 0.8667 & 0.9142 & 0.4518 & 0.2708 & 0.6310 & 40.6714 \\ \hline
        $64 \times 64$ & $30 \times N$ & 48.7131 & 0.9921 & 0.8086 & 0.8685 & 1.7496 & 2.3048 & 1.8696 & 32.0033 \\ \hline
        $128 \times 128$ & $30 \times N$ & 50.0089 & 0.9940 & 0.8278 & 0.9047 & 0.6520 & 0.3342 & 0.8361 & 39.2008 \\ \bottomrule
    \end{tabular}%
\end{table*}
\subsection{Robustness Analysis}
\textbf{Cross-dataset generalizability.} We evaluate generalizability across diverse distributions, including COCO2017 \cite{7lin2014microsoft2}, OpenForensics \cite{le2021openforensics8}, and CelebA-HQ \cite{karras2017progressive}. As reported in Table \ref{dataset}, DAWF demonstrates exceptional performance consistency.  Despite the varying facial characteristics and scene complexities of the three benchmarks, the  F1-score and AUC remain consistently high. This high degree of cross-dataset consistency, visually corroborated by the precise red masks on CelebA-HQ in Fig. \ref{celebahq}, strongly validates that DAWF's excellent generalizability and practical applicability in diverse environments. This confirms that our framework effectively learns a robust and generalized watermark embedding and extraction mechanism rather than overfitting to the specific biases of the training data. 

\textbf{Robustness under different levels of distortions.} As shown in Table \ref{tab:robustness_merged}, we conducted a robustness analysis comparing our DAWF and IML-Net \cite{ma2023iml} across varying distortion intensities. We observed that our method maintains high localization and bit accuracies, experiencing only a marginal performance drop. In contrast, IML-Net exhibits a substantial performance degradation compared to its results under clean conditions.

\subsection{Computational Cost}

As indicated in Table \ref{tab:model_time}, our embedding and extraction times are consistent with existing solutions. Crucially, while baseline models require a full forward pass to embed a single watermark, DAWF embeds nine distinct watermarks in parallel within the same timeframe. This makes our framework highly suitable for real-world, high-throughput deployment on resource-constrained devices.

\subsection{Ablation Study}
To verify the effectiveness of the core components in DAWF, we conducted ablation studies on the tracing loss ($\mathcal{L}_{tr}$), common distortion loss of detection and localization ($\mathcal{L}_{lo}^{com}$), and malicious distortion loss of detection and localization ($\mathcal{L}_{lo}^{mal}$), as summarized in Table \ref{tab:ablation_study}. The results demonstrate that $\mathcal{L}_{tr}$ is the prerequisite for establishing source tracing. Furthermore, the absence of $\mathcal{L}_{lo}^{com}$ causes the model to lose its immunity to benign edits, leading to significant false alarms. Notably, removing $\mathcal{L}_{lo}^{mal}$ causes a severe performance drop, with the F1-score dropping sharply from 0.8667 to 0.0007, confirming $\mathcal{L}_{lo}^{mal}$ as a crucial component for detection and localization. In contrast, our full model is clearly superior to the incomplete configurations in both robustness and precision.

Table \ref{tab:performance_metrics} evaluates the impact of image resolution and embedding payload on DAWF, revealing a clear trade-off: at a $64 \times 64$ resolution, increasing the payload from $15 \times N$ to $30 \times N$ degrades visual quality and increases the extraction BER. However, enlarging the input resolution to $128 \times 128$ effectively mitigates this high-payload degradation by providing a richer redundant feature space. This adjustment not only restores the PSNR to over 50 dB but also drastically reduces the Tracer's BER to below 0.5\%.

\subsection{IoU Threshold Selection}

\begin{figure}[t]
    \centering
    \includegraphics[width=0.48\textwidth]{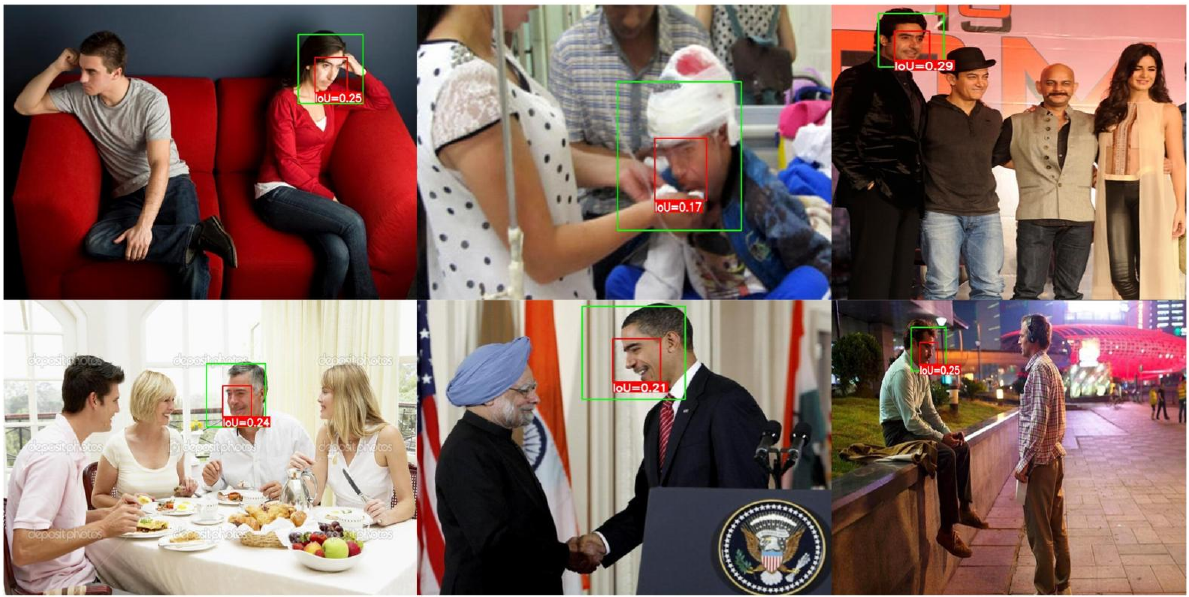}
    \caption{Visual samples of IoU matching results. The green boxes indicate ground-truth forged faces, while the red boxes represent the predicted forged faces.}
    \label{fig:tamper-region}
\end{figure}

\begin{figure}[t]
    \centering
    \includegraphics[width=0.45\textwidth]{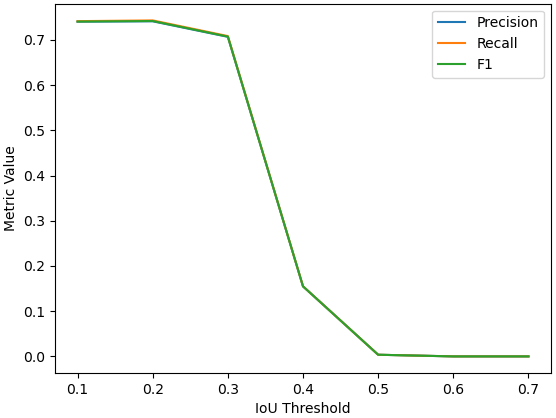}
    \caption{Quantitative analysis of forensic metrics across varying IoU thresholds. The metrics remain stable for $\tau \in [0.1, 0.3]$, but experience a sharp decline thereafter. This validates that the matching IoU values for forged faces are primarily concentrated between 0.3 and 0.4.}
    \label{fig:tamper-F1}
\end{figure}

Our framework employs the IoU threshold ($\tau$) strictly for face instance association. During training, it maps GT boxes to the detected watermark boxes to filter out authentic faces, ensuring the model focuses exclusively on manipulated face instances. During evaluation, we utilize the IoU value to determine whether the predicted face and the ground-truth forged face represent the same facial instance. As shown in Fig. \ref{fig:tamper-region}, the IoU between $\mathcal{F}_{pred}$ and $\mathcal{F}_{gt}$ ranges from 0.17 to 0.29. This low IoU range is caused by the boundary uncertainty of $\mathcal{F}_{gt}$, which is extracted from the malicious forgery noise layer, whereas $\mathcal{F}_{pred}$ corresponds to the $64 \times 64$ facial area. We studied different matching thresholds $\tau$. Quantitative results in Fig. \ref{fig:tamper-F1} show that F1-score remains near-optimal and stable within the $\tau \in [0.1, 0.3]$ range. This stability demonstrates that a 0.1 threshold does not cause ``identity crosstalk'', mistaking a neighboring face for the target. Using a conventional threshold of 0.5 falsely classifies correctly localized faces as false negatives. Consequently, $\tau = 0.1$ serves as a robust boundary for maintaining the integrity of the forensic pipeline within complex multi-face scenarios.

\subsection{Further Discussion}

While a 15-bit payload per facial instance is sufficient for most forensic requirements, DAWF can support a 30-bit payload capable of indexing over one billion distinct identities to accommodate larger-scale tracing demands. As demonstrated by the ablation study in Table \ref{tab:performance_metrics}, increasing the operational resolution to $128 \times 128$ enables DAWF to support a doubled payload of 30 bits while maintaining highly competitive performance. Nevertheless, DAWF is constrained by an inherent trade-off between payload capacity and local resolution. Specifically, embedding 15 bits into a $64 \times 64$ facial region yields a payload density of approximately 0.0037 bits per pixel, whereas embedding 128 bits into a $256 \times 256$ region amounts to approximately 0.0020 bits per pixel. This suggests that the smaller absolute payload of DAWF does not necessarily imply a more relaxed embedding configuration, given its substantially smaller operational region. Therefore, improving payload capacity for low-resolution area while preserving robustness and imperceptibility remains an important direction for future work.

\section{Conclusion}   
\label{sec:conclusion}
We pioneer the exploration of deepfake proactive forensics in complex multi-face scenarios, more closely reflecting real-world environments. In this paper, we propose the Deep Attributable Watermarking Framework (DAWF), which successfully unifies image-level detection, instance-level localization, and identity-level source tracing. By establishing an isolated identity attribution space, DAWF ensures that multiple independent tracing signals can coexist within a single image and be successfully anchored to their respective identity instances. Furthermore, by introducing a selective regional supervision loss to suppress cross-face interference, we guide the decoder to focus exclusively on forged facial regions, thereby achieving highly precise forgery detection and localization. Extensive experiments on the WIDERFace datasets demonstrate that DAWF achieves superior detection, localization, and tracing robustness. Concurrently, DAWF exhibits superior generalizability across diverse datasets, confirming its practical applicability for real-world deepfake forensics. Further work will concentrate on enhancing its generalization ability against complex distortions and unknown forgery technologies.

\bibliographystyle{IEEEtran}
\bibliography{sample-base}

\end{document}